\documentclass[letterpaper, 10 pt, conference]{ieeeconf}
\usepackage{times}
\usepackage{epsfig}
\usepackage{graphicx}
\usepackage{amsmath}
\usepackage{amssymb}
\usepackage{makecell}
\usepackage{booktabs}
\usepackage{subcaption}
\PassOptionsToPackage{hyphens}{url}\usepackage{hyperref}

\usepackage{multirow}

\begin{document}

\title{SFace: Privacy-friendly and Accurate Face Recognition using Synthetic Data}
%

\author{Fadi Boutros$^{1,2}$, Marco Huber$^{1,2}$, Patrick Siebke$^{1}$, Tim Rieber$^{1}$,
Naser Damer$^{1,2}$ \\
Fraunhofer Institute for Computer Graphics Research IGD,
Darmstadt, Germany\\
$^{2}$Mathematical and Applied Visual Computing, TU Darmstadt,
Darmstadt, Germany\\
 Email: {fadi.boutros@igd.fraunhofer.de}
}

\maketitle
\thispagestyle{empty}

\begin{abstract} 
Recent deep face recognition models proposed in the literature utilized large-scale public datasets such as MS-Celeb-1M and VGGFace2 for training very deep neural networks, achieving state-of-the-art performance on mainstream benchmarks.
Recently, many of these datasets, e.g., MS-Celeb-1M and VGGFace2, are retracted due to credible privacy and ethical concerns. This motivates this work to propose and investigate the feasibility of using a privacy-friendly synthetically generated face dataset to train face recognition models. 
Towards this end, we utilize a class-conditional generative adversarial network to generate class labeled synthetic face images, namely SFace. 
To address the privacy aspect of using such data to train a face recognition model, we provide extensive evaluation experiments on the identity relation between the synthetic dataset and the original authentic dataset used to train the generative model.
Our reported evaluation proved that associating
an identity of the authentic dataset to one with the same class label in the synthetic dataset is hardly possible. 
We also propose to train face recognition on our privacy-friendly dataset, SFace, using three different learning strategies, multi-class classification, label-free knowledge transfer, and combined learning of multi-class classification and knowledge transfer. 
The reported evaluation results on five authentic face benchmarks demonstrated that the privacy-friendly synthetic dataset has high potential to be used for training face recognition models, achieving, for example, a verification accuracy of 91.87\% on LFW using multi-class classification and 99.13\% using the combined learning strategy.
The training code and the synthetic face image dataset are publicly released \footnote{\url{https://github.com/fdbtrs/SFace-Privacy-friendly-and-Accurate-Face-Recognition-using-Synthetic-Data}}. 

\end{abstract}

\section{Introduction}
The advances in deep face recognition (FR) are mainly driven by the architecture, the used loss function, and the availability of large-scale training datasets. In the FR community, a serious discussion has arisen regarding many of the datasets in use, as critical challenges regarding the ethical and legal aspects of creating, using, and sharing, these datasets are increasingly being discussed. Many of the recent face image datasets that are used in the literature have been collected from the web \cite{DBLP:journals/corr/abs-2102-00813,DBLP:conf/fgr/CaoSXPZ18,DBLP:conf/eccv/GuoZHHG16} which might conflict with some of the privacy regulations that control the use of personal data without clear consent \cite{eu-2016/679}.

Processing biometric data is governed by a set of legal restrictions \cite{eu-2016/679,nec,bipa}. 
Taking the General Data Protection Regulation (GDPR) \cite{eu-2016/679} as an example, it categories biometric data as a special category of personal data subjected to rigorous data protection rules \cite{art9gdpr}, requiring high protection in connection with fundamental rights and freedoms of individuals.
Dealing with such data requires the adherence to one of the exemptions of biometric data processing \cite{art9_2gdpr}, adherence to the related national laws \cite{art9_4gdpr}, maintaining processing records \cite{art30gdpr}, and the preparation of data protection impact assessment \cite{art35_3gdpr,art37_1gdpr}, among other restrictions. 
Depending on the purpose of the biometric data processing, this set of restrictions can be rigorously extended \cite{art22_4gdpr,art27_2gdpr,art6_4gdpr}.
Besides the legal complications of using and sharing biometric data, ethical requirements are commonly necessary, such as the approval of an ethics committee or competent authorities.

A theoretically possible solution to enable the protection of personal data while being able to share biometric data is the personal data anonoymization. 
This is far from being a realistic solution given the high anonymization requirements set by authorities like the European Union \cite{Recital.26.GDPR}, in addition to the lack of data utility after anonoymization  \cite{DBLP:journals/tifs/MedenRTDKSRPS21}.
To overcome this issue in conventional anonymization techniques \cite{4497436,dwork.differentialPrivacy,Lindell_Pinkas_2009}, a more effective and data-driven technique is the use of synthetic data  \cite{9138552}, which aims at maintaining utility while permitting research continuation and the adherence to privacy requirements \cite{emam2020practical}.
Due to the legal privacy regulations briefly discussed above \cite{eu-2016/679}, collecting, using, sharing, and maintaining face image datasets for biometric processing may not be feasible currently and in the near future. 
Some widely-used databases, such as VGGFace2 \cite{DBLP:conf/fgr/CaoSXPZ18},
MS-Celeb-1M \cite{DBLP:conf/eccv/GuoZHHG16},
MegaFace \cite{kemelmacher2016megaface} or DukeMTMC \cite{DBLP:conf/eccv/RistaniSZCT16} are retracted by their creators based on increasing ethical and legal grounds.

This calls for solutions that take into account the privacy of individuals and the reproducibility and continuity of FR research, which requires shareable face image datasets. With the on-going success of Generative Adversarial Networks (GANs) in creating synthetic data for computer vision tasks (e.g. \cite{DBLP:conf/cvpr/WangGL019}), a new area of biometric research is emerging to question how synthetic face images can be created and used to train FR models. In comparison to other synthetic data generation challenges, this research faces the challenge of assigning identity in the synthetically generated faces to be utilizable, while additionally insure variations within identity.

\begin{figure*}[ht!]
    \centering
    \includegraphics[width=0.9\textwidth]{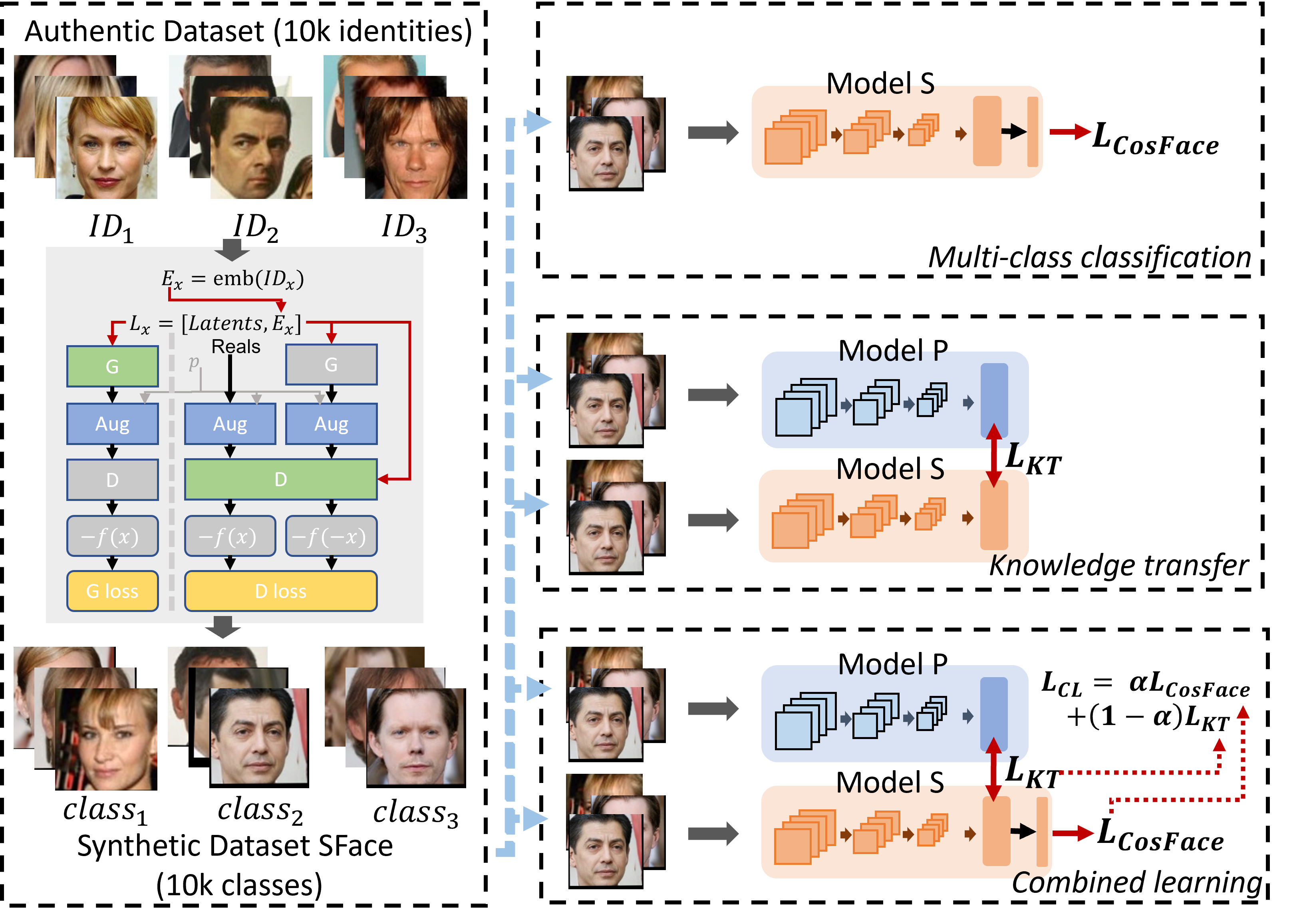}
    \caption{Overview of the proposed approach. On the left, a conditional StyleGAN2-ADA \cite{DBLP:conf/nips/KarrasAHLLA20} is used in combination with the identity labels to create our synthetic dataset SFace. The identity labels are first embedded and concatenated to the latent variables and then passed to the generator and discriminator of the GAN architecture. The three proposed learning strategies are shown on the right. In the multi-class classification loss learning strategy, the FR Model S is only trained on the synthetic face dataset SFace utilizing CosFace loss \cite{DBLP:conf/cvpr/WangWZJGZL018}. In the KT learning strategy, KT is utilized between the embedding layer of a pre-trained FR model P, trained on the authentic dataset, and the FR model S. In the combined learning strategy, both approaches are combined using a weighting factor $\alpha$. Note that all the three learning strategies do not require privacy-sensitive authentic data, the last two of the three require a pre-trained model, whose sharing does not require the share of personal data, along with the SFace data.}
    \label{fig:overview}
\end{figure*}

Motivated by the above-mentioned challenges, the use of synthetic data in biometrics has recently attracted attention. Zhang et al. \cite{DBLP:conf/iwbf/ZhangGRR021} investigated the behavior of face image quality assessment methods on synthetic images created by StyleGAN \cite{DBLP:conf/cvpr/KarrasLA19} and StyleGAN2 \cite{DBLP:conf/cvpr/KarrasLAHLA20} and compare the face image quality values with those of authentic face images. Another biometric use case based on synthetic data that has been studied so far is the creation of a morphing attack detection development dataset from synthetic data \cite{DBLP:journals/corr/abs-2203-06691}. Shen et al. \cite{DBLP:conf/fgr/ShenROBS21} investigated how human individuals perceive synthetically generated faces and showed that humans can be fooled by state-of-the-art synthetic face images. In the field of FR, SynFace \cite{DBLP:conf/iccv/QiuYG00T21} investigated the different behavior of FR models trained on authentic and synthetic images and proposed identity and domain mixup to reduce the performance gap of FR models trained on synthetic data in comparison to FR models trained on authentic data. 
Zhai et al. \cite{DBLP:conf/aaai/ZhaiYZHCYWP21} presented three approaches based on meta-learning, disentangling and filtering to reduce the modal difference between synthetic and authentic samples. Then, they combined synthetic with an authentic dataset to train a FR model.
Another approach in this direction was proposed by Trigueros et al. \cite{DBLP:journals/nn/Saez-TriguerosM21}. They disentangled identity-relevant attributes and non-identity-relevant attributes by training a GAN based on an identity latent space created by an embedding network to perform generative image augmentation.
In contrast, Kortylewski et al. \cite{DBLP:conf/cvpr/KortylewskiESGM19} did not use GANs but 3D models to generate synthetic images. These images were then used to reduce dataset bias in real-world datasets.

In this work, we contribute with a set of proposals towards training FR models without the need for authentic face data. 
First, to enable our proposed learning strategies, we created a synthetic face dataset by training a class-conditional GAN using the identities as class labels
Second, we propose to train FR using three different learning strategies based on synthetic images.
The classification loss learning strategy is based on a multi-class classification loss and is trained only with our synthetic images. The knowledge transfer learning strategy is based on using our synthetic data to transfer the information from a pre-trained FR model to train a FR model from scratch without the need for authentic data. 
In the combined learning strategy, the two mentioned learning strategies are combined to benefit from the classification optimization and the transferred knowledge, again, without the need for authentic data.
These contributions are accompanied by detailed novel analyses of the relation between the identities in the synthetic data and the authentic data used to train its generative model. 
This work also provides an analysis of the effect of the synthetic data size, in terms of samples per identity, on the performance of the trained FR model under the different learning strategies.
Our presented synthetic data and learning strategy proved, by a thorough evaluation on a number of FR benchmarks, to be successful in training FR models in a privacy-friendly framework.

\begin{figure*}[]
    \centering
    \includegraphics[width=0.7\textwidth]{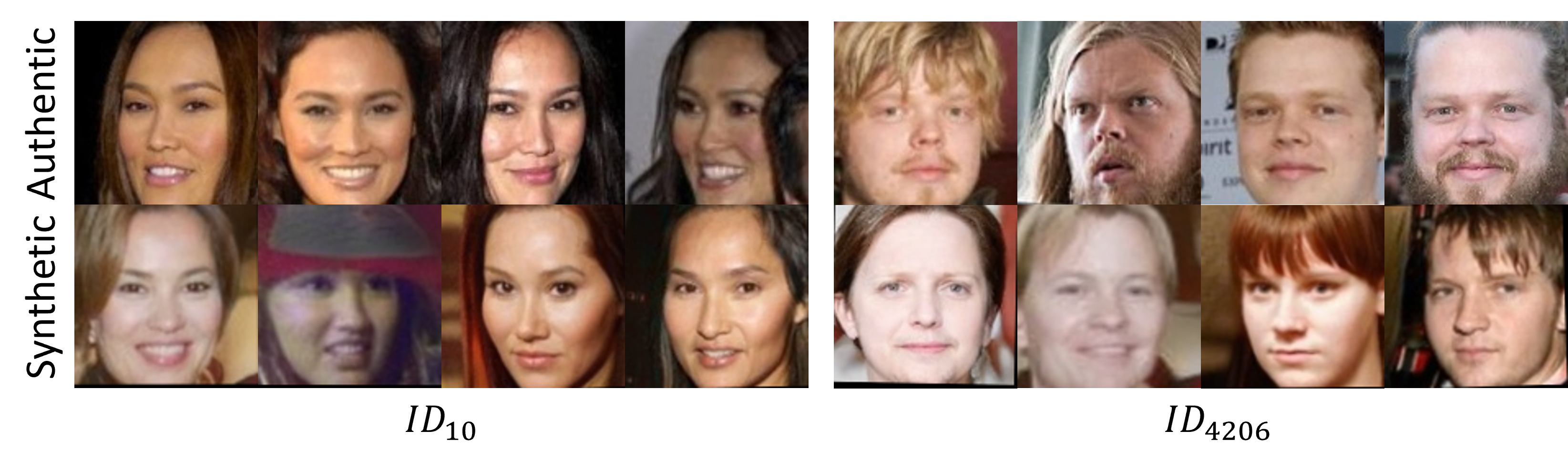}
    \caption{Examples from the authentic (CASIA-WebFace \cite{DBLP:journals/corr/YiLLL14a}) and SFace datasets. The synthetic images also show a large variance in appearance and pose. The images from $ID_{4206}$ also show that the generated images do not strongly correspond to the identity present in the authentic images but rather to the class.}
    \label{fig:samples}
\end{figure*}

\section{Synthetic Face Recognition} 
\label{sec:methodology}
In this section, we present our approach to generate and utilize synthetic face images to train a FR model that takes into account the privacy of the subjects depicted in authentic FR datasets. First, we train a conditional generative adversarial network (GAN) based on StyleGAN2-ADA \cite{DBLP:conf/nips/KarrasAHLLA20} using the authentic dataset and setting the identity labels as class labels to create synthetic data. We then propose three learning strategies to train FR models based on the generated synthetic face images.
An overview of our approach is shown in Figure \ref{fig:overview}. 
In the first learning strategy, the classification loss learning, we solely train the FR model on the synthetic data. The second learning strategy, the knowledge transfer (KT), the infromation is transferred (on embedding level) from a FR model (pre-trained on authentic images) to train a new model (from scratch) with the support of the generated synthetic data. Last, in the third learning strategy, the combined learning, we successfully propose to combine these two learning strategies. 

\subsection{Synthetic Data Generation}
This work utilizes a conditional GAN \cite{DBLP:conf/nips/KarrasAHLLA20,DBLP:conf/iclr/MiyatoK18} to synthetically generate face images. 
Specifically, we opt to use StyleGAN2-ADA for synthetic image generation. StyleGAN2-ADA \cite{DBLP:conf/nips/KarrasAHLLA20} is built on style-based GAN (StyleGAN2 \cite{DBLP:conf/cvpr/KarrasLAHLA20}) with adaptive discriminator augmentation (ADA) to increase the diversity of the training dataset, reducing the risk of overfitting the discriminator when the training data contains little samples. To generate class-conditional synthetic face images, we trained StyleGAN2-ADA \cite{DBLP:conf/nips/KarrasAHLLA20} under conditional settings, where the class labels in our training setting are the identity labels. Once the model is trained, we construct our synthetic face dataset (noted as SFace) by generating k images from each class $c \in C$ ($C$ is the number of classes). Examples of the data are shown in Figure \ref{fig:samples}.
Besides using this synthetic data to train FR within our learning strategies, the relation between the identities in the synthetic data and the data used to train its generator is later analyzed in section \ref{sec:result}.




\subsection{Learning Strategies}
To evaluate the feasibility of using synthetically generated face images for FR, we present three learning strategies to train FR models on a synthetically generated dataset. In the first learning strategy, we propose to train the FR model on the synthetic SFace dataset as a multi-class classification problem using conventional classification loss. In the second learning strategy, we propose to train the FR model on SFace with KT from a pre-trained FR model trained on real, authentic data. In the third learning strategy, we combine the first two approaches and propose to fuse the multi-class classification loss with KT into one loss function.

\paragraph{Multi-class Classification (CLS)}
In this learning strategy, we train the FR model solely on the synthetic SFace with the widely used CosFace loss \cite{DBLP:conf/cvpr/WangWZJGZL018} as multi-class classification loss. 
CosFace loss incorporates a margin penalty in the softmax loss to push its decision boundary and thus enhance intra-class compactness and inter-class discrepancy, aiming to improve FR accuracy. 
This learning strategy aims to guide the model to learn multi-class classification and then use the obtained network as a feature extractor for FR.
Formally, the CosFace is given by:
\begin{equation}
\label{eq:cosface}
\resizebox{\linewidth}{!}{$
    L_{CosFace}=\frac{1}{N}  \sum\limits_{i \in N} - log \frac{e^{s (cos(\theta_{y_i})-m)}}{ e^{s(cos(\theta_{y_i})-m)} +\sum\limits_{j=1 , j \ne y_i}^{c}  e^{s ( cos(\theta_{j}))}},
$}\end{equation}
where $m$ is the margin penalty applied on the cosine angle $cos(\theta_{y_i})$ between the feature embedding $x_i$ of training sample $i$ and its class center $y_i$.

\paragraph{Knowledge transfer (KT)}
In the second learning strategy, we propose to train an FR model (noted as S) on the SFace dataset with KT from the pretrained FR model (noted as P) trained on the authentic dataset. Different than the first learning strategy that incorporates only SFace dataset, this learning strategy requires SFace dataset and access to the P model, which is less privacy-critical than accessing personal biometric data. The main idea is to guide the model S to learn to produce feature representations similar to the ones produced by the model trained on an authentic dataset. 
Unlike the multi-class classification learning strategy, KT does not require class label supervision as the loss function is calculated between the feature representations of S and P models. 
During the training phase, a batch of SFace dataset is sampled and fed into both models, S and P, to obtain $f^S$ and $f^P$, respectively. Then, the mean squared error (noted as $L_{KT}$) is calculated between $f^S$ and $f^P$ as follows:
\begin{equation}
\label{eq:l2}
    L_{KT}= \frac{1}{N}  \sum\limits_{i \in N} \Big(  \frac{1}{d}\Sigma_{j=1}^{d}{\Big(F^S_{ij} -F^P_{ij}\Big)^2\Big)},
\end{equation}
where $d$ is the embedding layer size (the last layer before the classification layer) and $N$ is the batch size. As this learning strategy does not require class label supervision, it can be used to train FR on any synthetically generated data from unconditional face generative models.

\paragraph{Combined learning (CL)}
In the third learning strategy, we train the model to learn from the multi-class classification learning (Eq. \ref{eq:cosface}) and to produce a feature representation similar to one produced by the model trained on the authentic dataset, i.e. KT (Eq. \ref{eq:l2}). The combined loss, in this case, is given by:
\begin{equation}
    L_{CL} = \alpha L_{CosFace} + (1 - \alpha) L_{KT},
    \label{equ:cl}
\end{equation}
where $\alpha$ is a weighting parameter used to balance between the two training losses. 

Later in Section \ref{sec:result}, we analyze the identity relations between the authentic data and SFace, and the identifiability of the identities in the authentic data using the pre-trained model in comparison to their identifiability using our trained model, to point out the privacy-related enhancements induced by our solution.




\section{Experimental Setup}
This section presents the experimental setups and implementation details used in this work.
\subsection{Datasets and Benchmarks}
\label{sec:exp:data}
We used CASIA-WebFace \cite{DBLP:journals/corr/YiLLL14a} to train SyleGAN2-ADA and the base (pre-trained) FR model P, this data is noted as the "authentic" data. 
The dataset contains 494,414 images of 10,575 different identities. The images are aligned and cropped to $112 \times 112$ using landmarks obtained by the Multi-task Cascaded Convolutional Networks (MTCNN) \cite{zhang2016joint}, as described in \cite{deng2019arcface}. 
Our synthetic SFace dataset is constructed, as detailed in Section \ref{sec:methodology}, using StyleGAN2-ADA conditionally trained on CASIA-WebFace to generate 634K synthetic images (60 images per class) equally distributed on 10,575 classes.
Then, we derived four subsets from SFace noted as SFace-N: SFace-10, SFace-20, SFace-40, and SFace-60, where N is the number of images per class (10,575 classes) in each of the subsets.
SFace-10, SFace-20, SFace-40, and SFace-60 are derived by selecting the first 10, 20, 40, and 60 images of each identity of SFace.
We used SFace-10 (105K images ) to evaluate the identity separability within SFace and the identity relations between SFace identities and those of the authentic data (Casia-WebFace). 
SFace-20, SFace-40, and SFace-60 are used to investigate the evaluation performance of FR trained on different dataset sizes, along with SFace-10.
The images in SFace are aligned and cropped using the same approach used to preprocess the authentic data. 
The presented FR models in this work are evaluated on five benchmarks: Labeled Faces in the Wild (LFW)  \cite{LFWTech}, Cross-age LFW (CALFW)  \cite{DBLP:journals/corr/abs-1708-08197},  Cross-Pose LFW (CPLFW) \cite{CPLFWTech}, Celebrities in Frontal-Profile in the Wild (CFP-FP) \cite{DBLP:conf/wacv/SenguptaCCPCJ16} and AgeDB-30 \cite{DBLP:conf/cvpr/MoschoglouPSDKZ17}, each using the protocol defined by the benchmark and the respective performance metrics (verification accuracy in \%) defined in the benchmark.

\subsection{StyleGAN2-ADA Training Settings}
We trained StyleGAN2-ADA \cite{DBLP:conf/nips/KarrasAHLLA20} on the CASIA-WebFace dataset under conditional settings, i.e. generating images of specific classes. The number of classes is set to 10,575 (number of identities in CASIA-WebFace), which is embedded into a 512-D vector and then concatenated with the latent vector (512-D) to generate class-related synthetic images.
We kept most of the implementation settings of StyleGAN2-ADA unchanged (as set in \cite{DBLP:conf/nips/KarrasAHLLA20}), including the training setup, network architectures, optimizer, and training loss. We have only set the StyleGAN2-ADA model training epochs to 50, the batch-size to 128, and the learning rate to $0.0025$.

\subsection{Face Recognition Training Setups}
The presented FR models in this paper utilize ResNet-50 \cite{DBLP:conf/cvpr/HeZRS16} as a backbone architecture.
We train five instances of ResNet-50. The first instance is trained with CosFace loss \cite{DBLP:conf/cvpr/WangWZJGZL018} on the authentic CASIA-WebFace dataset \cite{DBLP:journals/corr/YiLLL14a} and is used for KT (model P). 
We followed the parameter selection in \cite{DBLP:conf/cvpr/WangWZJGZL018} for the CosFace loss margin value, and scale parameter to 0.35 and 64, respectively. 
The other four instances are trained on the synthetic SFace dataset. The second instance is solely trained with CosFace loss, i.e. CLS, on the SFace dataset. The third instance is trained with only KT loss. The fourth and the fifth instances are trained with the CL loss with $\alpha$ in Equation \ref{equ:cl} of 1e-5 and 2e-5, respectively. The models trained on synthetic data are always noted with the training subset (SFace-10, SFace-20, SFace-40, or SFace-60) and the training strategy (CLS, KT, or CL). 

The presented models in this paper are implemented using Pytorch \cite{NEURIPS2019_9015}.
For all considered FR models, we set the mini-batch size to 512 and train the FR models with Stochastic Gradient Descent (SGD) optimizer.
We set the momentum to 0.9 and the weight decay to 5e-4. 
The FR models trained on the synthetic SFace required 64 epochs to converge and the one trained on authentic data required 32 epochs to converge. 
The initial learning rate is set 1e-1 and divided by 10 at 40, 48, and 52 epochs for the models trained on SFace and divided by 10 at 20 and 28 epochs for the model trained on the authentic dataset. 
During the training, we used random horizontal flipping with a probability of 0.5 for data augmentation. All training and testing images are normalized to have pixel values between -1 and 1.

\begin{table}[]
\centering
\resizebox{\linewidth}{!}{%
\begin{tabular}{llllll}
\hline
                        & EER  (\%)  & FMR100 (\%)  & FMR1000 (\%)  \\ \hline
SFace                   & 21.650 & 62.275 & 83.962        \\
CASIA-WebFace           & 7.614  & 9.152  & 10.712      \\
CASIA-WebFace vs. SFace & 29.204 & 74.662 & 88.692       \\ \hline
\end{tabular}%
}
\caption{Verification performances reported on three evaluation settings, SFace-10, CASIA-WebFace, and cross-dataset (reference images are taken from CASIA-WebFace and probe images are taken from SFace-10). 
The low verification performance in the cross-dataset evaluation is observable, proving that SFace-10 and CASIA-WebFace identities are hardly associated.}
\label{tab:verf}
\end{table}

%

\begin{figure*}[ht!]
     \centering
     \begin{subfigure}[b]{0.24\linewidth}
         \centering
         \includegraphics[width=\linewidth]{
         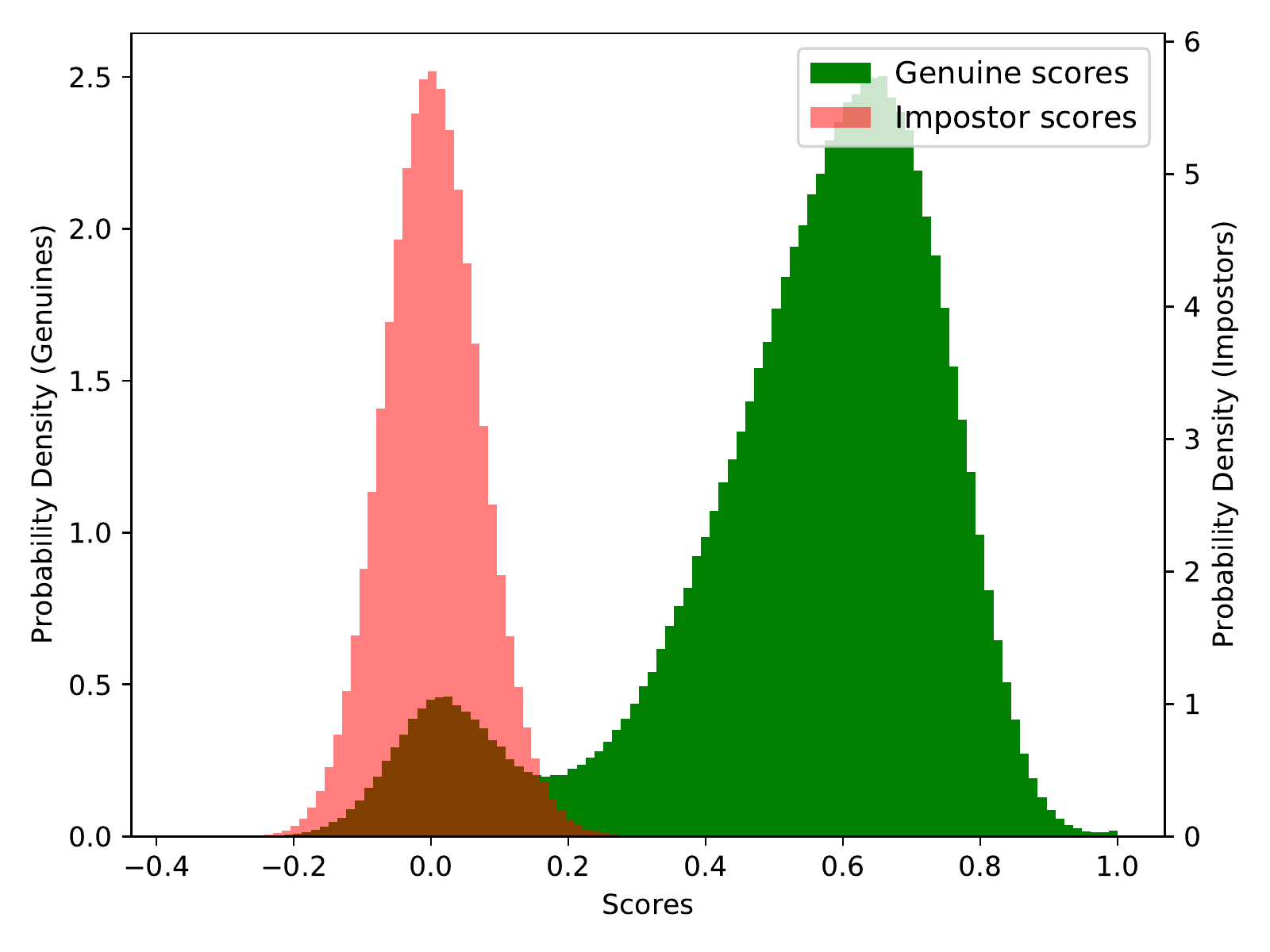}
         \caption{CASIA-WebFace
         \newline
         }
         \label{fig:dist_casia}
     \end{subfigure}
     \begin{subfigure}[b]{0.24\linewidth}
         \centering
         \includegraphics[width=\linewidth]{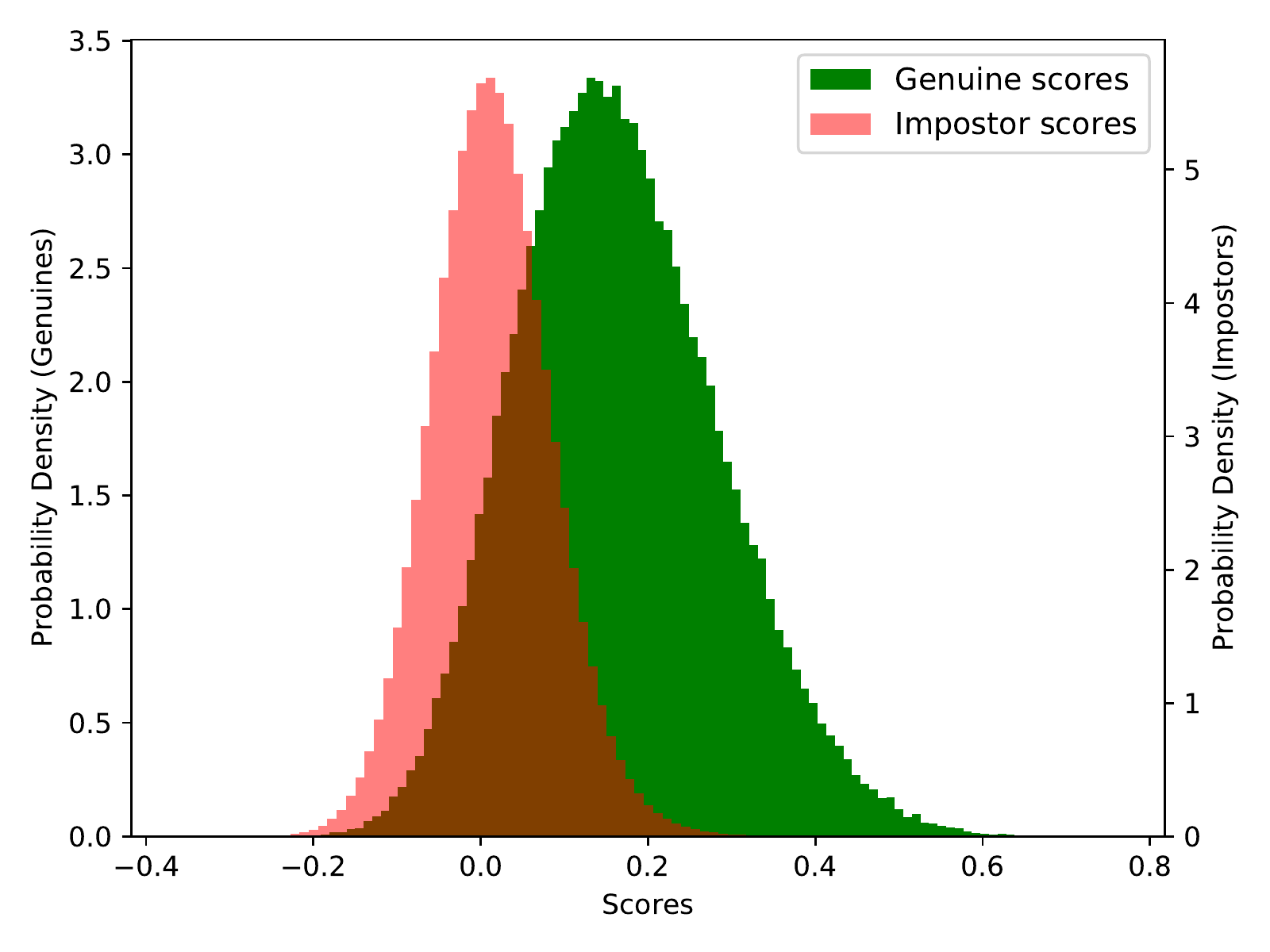}
         \caption{SFace-10
          \newline
          }
         \label{fig:dist_sface}
     \end{subfigure}
     \begin{subfigure}[b]{0.24\linewidth}
         \centering
         \includegraphics[width=\linewidth]{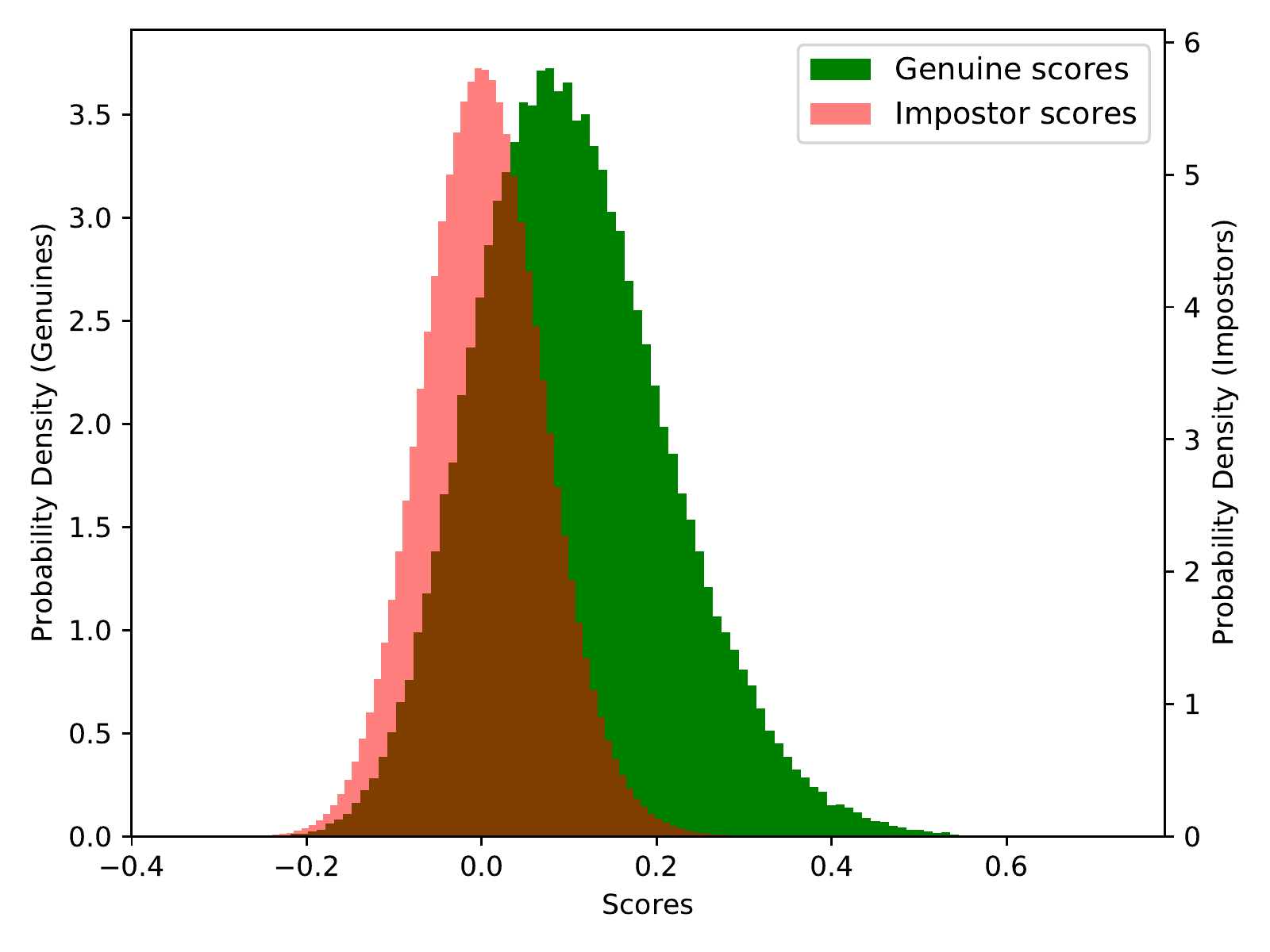}
         \caption{
         CASIA-WebFace - SFace-10
                 \\\hspace{\textwidth}}
         \label{fig:dist_casia_sface}
     \end{subfigure}
          \begin{subfigure}[b]{0.24\linewidth}
         \centering
         \includegraphics[width=\linewidth]{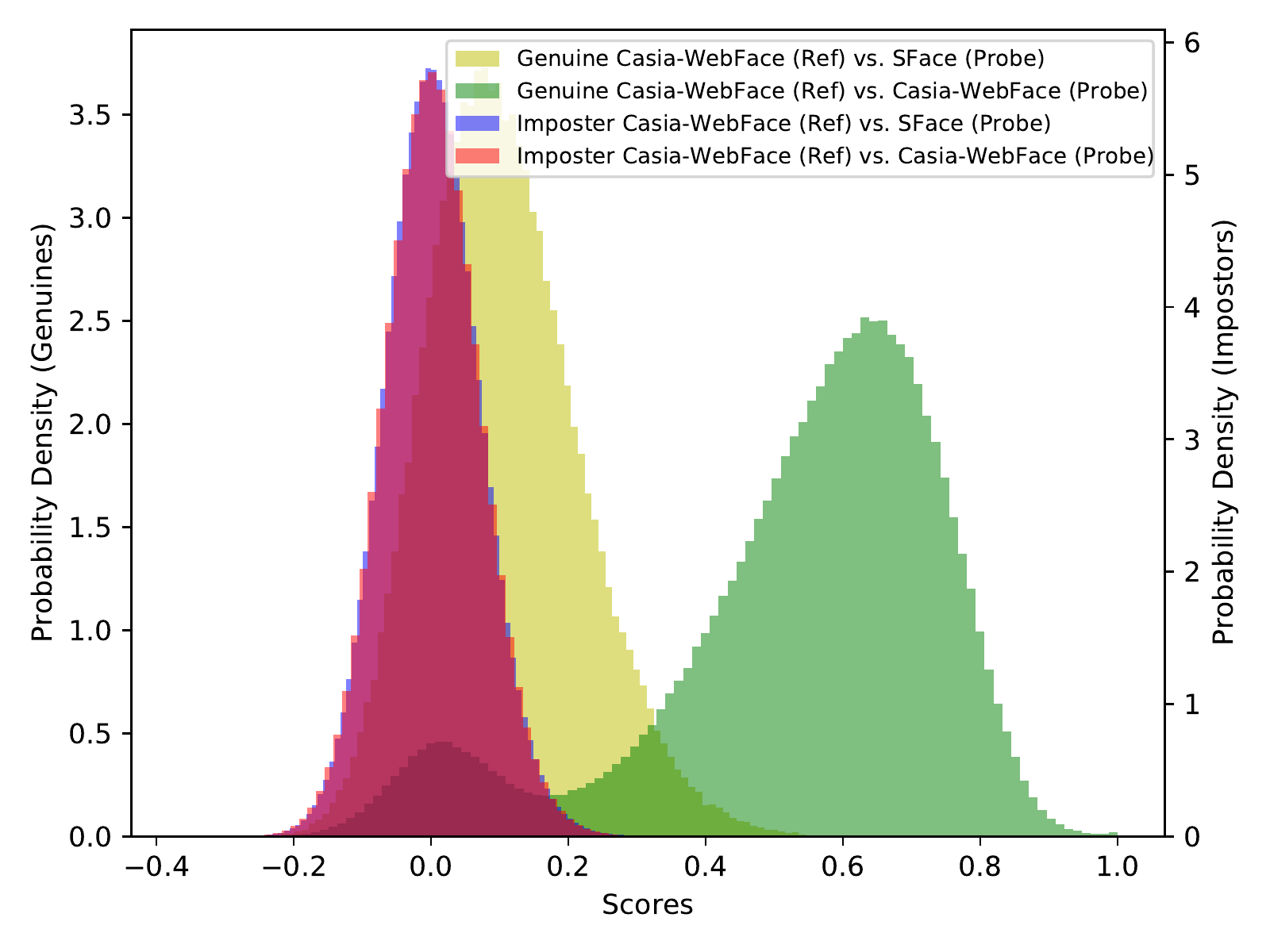}
         \caption{CASIA-WebFace - SFace-10 vs. Casia-WebFace }
         \label{fig:dist_casia_sface_casia}
     \end{subfigure}
        \caption{ 
        The genuine and imposter score distributions of different experimental settings. The genuine score distribution is overlapping with the imposter score distribution in the SFace-10 (Figure. \ref{fig:dist_sface}) in comparison to the authentic data in \ref{fig:dist_casia}. However, this overlap is lower than that in the cross-dataset evaluation (Figure. \ref{fig:dist_casia_sface} and \ref{fig:dist_casia_sface_casia}), indicating weak identity link between the authentic data and the SFace data. The comparison scores are based on the cosine similarity calculated between the embeddings extracted by \cite{DBLP:journals/corr/abs-2109-09416} as described in Section \ref{sec:evMet}.
       }
        \label{fig:distribution}
\end{figure*}

\subsection{Evaluation Metrics}
\label{sec:evMet}
In this work, we investigate the identity-separability in the synthetic SFace dataset, and the identity-relation between SFace and the authentic CASIA-WebFace dataset \cite{DBLP:journals/corr/YiLLL14a} used to train the generative model. To achieve that, we report the verification performances of SFace, CASIA-WebFace, and cross-evaluation setting where the references are from Casia-WebFace and the probes are from SFace, using the conditional classes used for the StyleGan2-ADA as an identity label.
As SFace and Casia-WebFace, do not contain pre-defined comparison pairs, we considered the first two images of each identity in each dataset as a reference, and the rest are considered probes.
The verification performance is reported as  
FMR100, and FMR1000, which are the lowest false non-match rate (FNMR) for a false match rate (FMR)$\leq$1.0\% and $\leq$0.1\%, respectively, along with the Equal Error Rate (EER) representing FMR or FNMR at the decision threshold where they are equal.   
The verification performances are indicated by plotting the histogram of genuine and imposter score distribution plots. We used a pre-trained ElasticFace \cite{DBLP:journals/corr/abs-2109-09416} model \footnote{The network architecture of ElasticFace-Arc (CVPRW2022 \cite{DBLP:journals/corr/abs-2109-09416}) is ResNet100 trained on MS1MV2 \cite{DBLP:conf/eccv/GuoZHHG16} by ElasticFace authors (model publicly available).} to extract the feature embeddings of SFace and CASIA-WebFace for the investigations describe in this subsection.

The resulting FR models trained on our SFace dataset are evaluated on a set of widely used face recognition benchmarks and the reported performances follow the benchmarks defined evaluation protocols and metrics, all mentioned earlier in Section \ref{sec:exp:data}.

\begin{table*}[ht!]
\centering
\begin{tabular}{ll|llll}
\hline
\multicolumn{2}{l|}{Model / Training dataset} &
  \begin{tabular}[c]{@{}l@{}}CLS/\\ CASIA-WebFace\end{tabular} &
  \begin{tabular}[c]{@{}l@{}}CLS/\\ SFace-10\end{tabular} &
  \begin{tabular}[c]{@{}l@{}}CL ($\alpha=2e-5$)/\\ SFace-10\end{tabular} &
  \begin{tabular}[c]{@{}l@{}}CL($\alpha=1e-5$)/\\ SFace-10\end{tabular} \\ \hline
\multicolumn{1}{l|}{\multirow{2}{*}{Testing dataset}} &
  SFace-10 &
  0.06 &
  99.99 &
  99.28 &
  94.18 \\
\multicolumn{1}{l|}{} &
  CASIA-WebFace &
  92.69 &
  0.04 &
  0.14 &
  0.22 \\ \hline
\end{tabular}%
\caption{Top-1 identification accuracy (\%) of four FR models trained under different experimental settings. 
The identification accuracy is very low when testing on a different dataset than the one used for training, pointing out that the authentic dataset and the synthetic dataset (SFace) do not strongly share identity information. }
\label{tab:identification}
\end{table*}

\section{Results}
\label{sec:result}

In this section we present our findings under a set of research questions leading to the final FR performance using our SFace dataset and the presented learning strategies. 

\paragraph{Does a generative model trained under a class-conditional setting generate identity-separable face image classes?}
Figure \ref{fig:dist_sface} shows the genuine-imposter comparison score distributions on the synthetic SFace-10 dataset, where both references and probes images are from SFace. It can be observed that the genuine and imposter distributions are, to some degree, shifted from each other.  The genuine score distribution in Figure  \ref{fig:dist_sface} is shifted towards the imposter score distributions in comparison to the case where the references and probes images are from the authentic CASIA-WebFace dataset shown in Figure \ref{fig:dist_casia}.  
Related to that, one can notice a degree of shared visual attributes in the SFace images from the same class in Figure \ref{fig:samples}.
These observations are also confirmed quantitatively in Table \ref{tab:verf}, where in comparison to the relatively strong identity discrimination in the authentic data (EER=7.614\%), the SFace data does maintain a certain degree of identity discrimination (EER=21.650\%), however, to a lower degree than the authentic data.



\paragraph{Does SFace share identity information with CASIA-WebFace?}
We answered this question by conducting a cross-dataset FR evaluation where authentic reference images from CASIA-WebFace are compared to synthetic probe images from SFace-10 using  \cite{DBLP:journals/corr/abs-2109-09416} as described in Section \ref{sec:evMet}. Figure \ref{fig:dist_casia_sface} shows the genuine and imposter scores distributions of the cross-dataset evaluation.
One can clearly notice that the genuine score distribution is highly overlapping with the imposter score distribution (Figure \ref{fig:dist_casia_sface}). This case is further illustrated by plotting the genuine and imposter score distributions of intra-data CASIA-WebFace and the cross-dataset settings in the same figure (Figure \ref{fig:dist_casia_sface_casia}), where the cross-data verification produces extremely lower genuine-imposter scores separability.   
Also, the reported verification performance in Table \ref{tab:verf} showed that the verification performance is significantly decreased when the probe images are taken from SFace (29.204\% EER ) compared to the case where both, reference and probe images, are from CASIA-WebFace (7.614\% EER).
This also indicates that associating an identity of SFace to one of Casia-WebFace confidently is hardly possible (29.204\% EER, 74.662\% FMR100, and 88.692\% FMR1000).
This practically means that, if operating at the FMR1000 threshold recommended by Frontex \cite{frontex2015best} for border control, an authentic image included in CASIA-WebFace, if compared to a 100 images of the same class in SFace using a top-performing FR solution \cite{DBLP:journals/corr/abs-2109-09416}, will result in a (false) non-match decisions around 89 times, rendering the verification of the authentic identities in SFace extremely weak, which support the privacy motivation behind the SFace synthetic data.

To complement our conclusion, we empirically evaluated how well can the classes of SFace (SFace-10) be correctly classified as the corresponding identities in CASIA-WebFace. 
We thus measure this correct classification ability as top-1 identification accuracy (\%) by calculating the average prediction accuracy from the classification layer.
Towards that, as a baseline, we first trained an FR model on CASIA-WebFace (CLS/CASIA-WebFace setting in Table \ref{tab:identification}) and then reported the identification accuracies (\%) on CASIA-WebFace and SFace. 
Table \ref{tab:identification} proved that images in CASIA-WebFace can be strongly identified (92.69\% accuracy). On the other hand, the identification accuracy of SFace in the model trained on Casia-WebFace is extremely low (0.06\%),  indicating that SFace classes have an extremely low association with the identities in Casia-WebFace.
This is expected based on, and supports, the results presented in Table \ref{tab:verf} and discussed above.
Similar observation can be made by training FR models on SFace under different learning strategies and evaluating the identification accuracies on SFace and CASIA-WebFace.
It can be noticed that the correct identification rate of the authentic data is extremely low in the models trained on SFace, even in the case where knowledge is transferred from the model trained on CASIA-WebFace (CL($\alpha=2e-5$) / SFace-10 and CL($\alpha=1e-5$) / SFace-10).
This low identification accuracy in comparison to that in the model trained on the authentic data (0.04\%, 0.14\%, 0.22\% in comparison to 92.69\%) again points out the low class association between the authentic data and the synthetic SFace data.

\begin{table*}[ht!]
\centering
\begin{tabular}{lllllll}
\hline
Training dataset         & Training strategy & LFW   & CFP-FP & AgeDB-30 & CALFW & CPLFW \\ \hline
Authentic & CLS               & 99.55 & 95.31  & 94.55    & 93.78 & 89.95 \\ \hline \hline
SFace-10        & CLS               & 87.13 & 68.84  & 63.30    & 73.47 & 66.82 \\
SFace-10        & KT                & 91.32  & 67.81   & 69.72      & 78.03  & 67.40 \\
SFace-10        & CL($\alpha$=2e-5)        & 94.75 & 75.20  & 77.28    & 84.15 & 73.32 \\
SFace-10 &
  CL($\alpha$=1e-5) &
  \textbf{96.63} &
  \textbf{79.61} &
  \textbf{82.08} &
  \textbf{87.43} &
\textbf{76.23} \\ \hline
\end{tabular}%
\caption{
Verification accuracy in \% of different learning strategies and different training datasets on five different FR benchmarks. The result in the first row is reported using the FR model trained on the authentic dataset to give an indication of the performance of an FR model trained on the authentic CASIA-WebFace \cite{DBLP:journals/corr/YiLLL14a} dataset. The rest of the results are obtained using the different proposed learning strategies on SFace-10. KT achieved very close performance to CLS without class label supervision. CL achieved the best verification performances with $\alpha =1e-5$.
}
\label{tab:sface}
\end{table*}

\paragraph{Can the synthetic SFace dataset be used to train an FR model?}
Knowing that the identities are, to a certain degree, separable in SFace, we evaluated the feasibility of using this synthetic dataset to train an FR model under the three different learning strategies presented in Section \ref{sec:methodology}.

In the first learning strategy, we trained an FR model to learn multi-class classification utilizing CosFace loss \cite{DBLP:conf/cvpr/WangWZJGZL018} (i.e. CLS).
The achieved verification accuracies on the considered evaluation dataset, in this case (CLS with SFace-10), are LFW (86.98\%), CFP-FP (68.51\%), AgeDB-30 (62.40\%), CALFW(73.10\%) and CPLFW(66.55\%), as shown in Table \ref{tab:sface}.
These verification accuracies improved when using SFace-20, SFace-40, or SFace-60 \ref{tab:sface_all}.  For example, the achieved accuracy on CALFW is improved from 73.47\% (SFace-10) to 77.93\% (SFace-60), as demonstrated in Table \ref{tab:sface_all}.
Also, it can be noticed that the verification accuracies of models trained with CLS on SFace-20, SFace-40, and SFace-60 are, in general, very close, when compared to the performance leap achieved when moving from SFace-10 to SFace-20. 
The results on all benchmarks are far away from being random, indicating the high potential of using synthetically generated data under class-conditional settings for FR.

In the second learning strategy, we trained FR models on the synthetic SFace dataset with KT.
This learning strategy does not require class label supervision.
Using SFace-10, it achieved close performance to the case where the model was trained using the CLS strategy. However, the achieved verification accuracies are significantly improved in SFace-20, SFace-40, and SFace-60 settings as shown in Table \ref{tab:sface_all}. The model trained with our SFace-60 under the KT strategy achieved an accuracy of 98.5\% on the LFW benchmark. 

As the next experiments showed, these results are significantly improved when both learning strategies are combined.
First, we set $\alpha$ to a very small value, as the previous results on CLS and KT strategies showed the higher performance when using KT, and we trained an FR model on the SFace-10 dataset using CL. The achieved results, in this case, are presented in Table \ref{tab:sface}.
Then, we increased $\alpha$ by a factor of 2 (2e-5) and trained another FR instance. We observed, in this case, that using a smaller $\alpha$ value led to higher verification performance on the considered benchmarks, as shown in Table \ref{tab:sface}. 
Thus, we set $\alpha$ to 1e-5 in the conduced experiments presented in Table \ref{tab:sface_all}.
The achieved results of CL using different subsets of SFace are presented in Table \ref{tab:sface}.
It can be observed, from the results in TAble \ref{tab:sface_all}, that combining CLS and KT (into CL) learning can improve the verification performance very significantly in comparison to the case where only CLS is used. For example, comparing the performance of the CLS strategy to that of the CL strategy when trained on SFace-60, the accuracy on LFW moved from 91.87\% (CLS) to 99.13\% (CL) and the accuracy on CALFW moved from 77.93\% (CLS) to 92.47\%. One can also notice that the achieved performance using the CL strategy is quite close to the performance of the model trained on the authentic data.

Two main previous works utilized synthetic data for FR training \cite{DBLP:conf/iccv/QiuYG00T21} and \cite{DBLP:conf/cvpr/KortylewskiESGM19}.  The presented solution (SynFace) by Qiu et al.  \cite{DBLP:conf/iccv/QiuYG00T21} (ICCV2021) reported accuracy of 88.98\% on LFW using a ResNet-50 model trained on synthetic dataset (500K images). The used  synthetic dataset by \cite{DBLP:conf/iccv/QiuYG00T21} consists of 10K different identities with 50 images per identity.
The work by Kortylewski et al. \cite{DBLP:conf/cvpr/KortylewskiESGM19}(CVPRW2019) reported verification accuracy of 80.1\% on LFW using FaceNet \cite{DBLP:conf/cvpr/SchroffKP15} trained on a synthetically generated dataset. Both previous works are outperformed by all of our three learning strategies achieving an LFW accuracy of 91.87\%, 98.50\%, and 99.13\% for the CLS, KT, and CL strategies, respectively.

To conclude, this work is one of the earliest works proposing the use of a synthetically generated face dataset for FR training. This is performed with and without the support of a pre-trained model. Also, to the best of our knowledge, it is the first work to analyze the identity relation between the synthetically generated face dataset and the authentic data used to train its generative model. 



\begin{table*}[]
\centering

\begin{tabular}{lllllll}
\hline
Training dataset & Training strategy    & LFW   & CFP-FP & AgeDB-30 & CALFW & CPLFW \\ \hline
Authentic        & CLS                  & 99.55 & 95.31  & 94.55    & 93.78 & 89.95 \\ \hline \hline
SFace-10         & \multirow{4}{*}{CLS} & 87.13 & 68.84  & 63.30    & 73.47 & 66.82 \\
SFace-20         &                      & 90.50 & 73.33  & 69.17    & 76.35 & 71.17 \\
SFace-40         &                      & 91.43 & 73.10  & 69.87    & 76.92 & 73.42 \\
SFace-60         &                      & 91.87 & 73.86  & 71.68    & 77.93 & 73.20 \\ \hline \hline
SFace-10         & \multirow{4}{*}{KT}  & 91.32 & 67.81  & 69.72    & 78.03 & 67.40 \\
SFace-20         &                      & 97.13 & 81.80  & 85.35    & 88.32 & 77.90 \\
SFace-40         &                      & 98.25 & 87.10  & 88.23    & 89.97 & 81.98 \\
SFace-60         &                      & 98.50 & 87.70  & 89.45    & 90.98 & 82.60 \\ \hline \hline
SFace-10         & \multirow{4}{*}{CL}  & 96.63 & 79.61  & 82.08    & 87.43 & 76.23 \\
SFace-20         &                      & 98.70 & 88.00  & 87.87    & 90.62 & 83.38 \\
SFace-40         &                      & 99.10 & 90.41  & 90.27    & 92.05 & 86.35 \\
SFace-60         &                      & 99.13 & 91.14  & 91.03    & 92.47 & 87.03 \\ \hline
\end{tabular}%
\caption{Verification accuracies in \% of different learning strategies and different subsets of SFace on the considered FR evaluation benchmarks. The $\alpha$ hyper-parameter is set to 1e-5 as it led to best performances (Table \ref{tab:verf}). One can notice that increasing the synthetic training data size in terms of samples per identity (going from SFace-10 to SFace-60) does generally increase the FR performance, however with some saturation behavior when moving to SFace-40 and SFace-60.   }
\label{tab:sface_all}
\end{table*}

\section{Conclusion}
This work is motivated by the need for privacy-friendly biometric data to enable further development of FR, where many privacy, legal, and ethical concerns are raised regarding the collection, use, and sharing of authentic biometric data. 
Towards that, we first created a synthetic face dataset, SFace, by training a StyleGAN2-ADA under class-conditional settings, and then, using the generative model to generate our synthetic data, the SFace.
Using SFace, we provided intensive evaluation experiments to address three main research questions posed in this paper regarding the identity separability in the synthetic SFace dataset, the identity-relation between the synthetic dataset and authentic dataset used to train its generator, and the feasibility of using synthetic data to train FR under different presented learning strategies. 
Our investigation proved that SFace possesses, to a certain degree, the identity discriminant information, the identity-relation between SFace and the authentic datasets is extremely weak, and SFace achieved relatively high verification performances on a wide set of benchmarks using the presented learning strategies.

{\small
\bibliographystyle{ieee}
\bibliography{output}
}

\end{document}